\pdfoutput=1

\documentclass[11pt,a4paper]{article}
\usepackage[hyperref]{acl2021}
\usepackage{times}
\usepackage{latexsym}

\usepackage{microtype}

\aclfinalcopy 

\usepackage{booktabs, colortbl} 
\usepackage{multirow,makecell} 
\usepackage{multicol}
\usepackage{longtable}
\usepackage{amsmath}
\usepackage{amssymb}
\usepackage{caption}
\usepackage{subcaption}
\usepackage{titlesec}
\usepackage{pgfplots}
\usepackage[normalem]{ulem} 
\usepackage{enumitem}
\usepackage{graphicx} 

\colorlet{LightGray}{gray!40}

\title{Improving Zero-Shot Translation by Disentangling \\Positional Information}

 \author{
        Danni Liu$^1$, Jan Niehues$^1$, James Cross$^2$, Francisco Guzm\'{a}n$^2$, Xian Li$^2$ \\
        $^1$Department of Data Science and Knowledge Engineering, Maastricht University \\ 
        $^2$Facebook AI\\ 
        \small{\texttt{\{danni.liu,jan.niehues\}@maastrichtuniversity.nl}} \\
        \small{\texttt{\{jcross,fguzman,xianl\}@fb.com }}
        }

\date{}

\begin{document}

\maketitle

\begin{abstract}
Multilingual neural machine translation has shown the capability of directly translating between language pairs unseen in training, i.e.  zero-shot translation.
Despite being conceptually attractive, it often suffers from low output quality.
The difficulty of generalizing to new translation directions suggests the model representations are highly specific to those language pairs seen in training.
We demonstrate that a main factor causing the language-specific representations is the positional correspondence to input tokens.
We show that this can be easily alleviated by removing residual connections in an encoder layer.
With this modification, we gain up to 18.5 BLEU points on zero-shot translation while retaining quality on supervised directions.
The improvements are particularly prominent between related languages, where our proposed model outperforms pivot-based translation.
Moreover, our approach allows easy integration of new languages, which substantially expands translation coverage.
By thorough inspections of the hidden layer outputs, we show that our approach indeed leads to more language-independent representations.\footnote{Code and scripts available in: \url{https://github.com/nlp-dke/NMTGMinor/tree/master/recipes/zero-shot}}
\end{abstract}

\section{Introduction}
Multilingual neural machine translation (NMT) system encapsulates several translation directions in a single model \cite{firat2017multi,johnson-etal-2017-googles}.
These multilingual models have been shown to be capable of directly translating between language pairs unseen in training \cite{johnson-etal-2017-googles, ha2016toward}.
Zero-shot translation as such is attractive both practically and theoretically.
Compared to pivoting via an intermediate language, the direct translation halves inference-time computation and circumvents error propagation.
Considering data collection, zero-shot translation does not require parallel data for a potentially quadratic number of language pairs, which is sometimes impractical to acquire especially between low-resource languages.
Using less supervised data in turn reduces training time.
From a modeling perspective, zero-shot translation calls for language-agnostic representations, which are likely more robust and can benefit low-resource translation directions.

\begin{figure}[t] 
\centering
    \centering
    \includegraphics[width=0.4\textwidth]{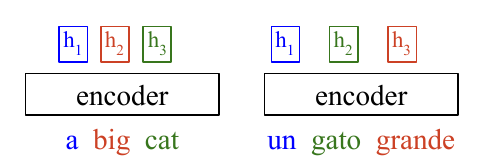}
\caption{\label{fig:pos} An example of language-specific encoder outputs as a results of the strong positional correspondence to input tokens (even assuming the word embeddings are cross-lingually mapped).}
\end{figure}

Despite the potential benefits, achieving high-quality zero-shot translation is a challenging task.
Prior works \cite{arivazhagan2019missing, zhang-etal-2020-improving, rios-etal-2020-subword} have shown that standard systems tend to generate poor outputs, sometimes in an incorrect target language.
It has been further shown that the encoder-decoder model captures spurious correlations between language pairs with supervised data  \cite{gu-etal-2019-improved}.
During training, the model only learns to encode the inputs in a form that facilitates translating the supervised directions.
The decoder, when prompted for zero-shot translation to a different target language, has to handle inputs distributed differently from what was seen in training, which inevitably degrades performance.
Ideally, the decoder could translate into any target language it was trained on given an encoded representation independent of input languages.
In practice, however, achieving a language-agnostic encoder is not straightforward.

In a typical Transformer encoder \cite{transformer}, the output has a strong positional correspondence to input tokens.
For example in the English sentence in Figure \ref{fig:pos},  encoder outputs $\text{h}_{1, 2, 3}$ correspond to ``\textit{a}'', ``\textit{big}'', ``\textit{cat}'' respectively.
While this property is essential for tasks such as sequence tagging, it hinders the creation of language-independent representations.
Even assuming that the input embeddings were fully mapped on a lexical level (e.g. ``\textit{cat}'' and ``\textit{gato}'' have the same embedding vector),
the resulting encoder outputs are still language-specific due to the word order differences.
In this light, we propose to relax this structural constraint and offer the model some freedom of word reordering in the encoder already. 
Our contributions are as follow:
\begin{itemize}
\item We show that the positional correspondence to input tokens hinders zero-shot translation. 
We achieve considerable gains on zero-shot translation quality by only removing residual connections once in a middle encoder layer.
\item Our proposed model allows easy integration of new languages, which enables zero-shot translation between the new language and all other languages previously trained on.
\item Based on a detailed analysis of the model's intermediate outputs, we show that our approach creates more language-independent representations both on the token and sentence level.
\end{itemize}

\section{Disentangling Positional Information} \label{sec:approach}

Zero-shot inference relies on a model's generalizability to conditions unseen in training.
In the context of zero-shot translation, the input should ideally be encoded into an language-agnostic representation, based on which the decoder can translate into any target language required, similar to the notion of an interlingua.
Nevertheless, the ideal of ``any input language, same representation'' cannot be easily fulfilled with a standard encoder, as we have shown in the motivating example in Figure \ref{fig:pos}. 

We observe that the encoder output has a positional correspondence to input tokens.
Formally, given input token embeddings $(\mathbf{x}_1, \dots, \mathbf{x}_n)$, in the encoder output $(\mathbf{h}_1, \dots, \mathbf{h}_n)$, the $i$-th hidden state $\mathbf{h}_i$ mostly contains information about $\mathbf{x}_i$.
While this structure is prevalent and is indeed necessary in many tasks such as contextual embedding and sequence tagging, it is less suitable when considering language-agnostic representations.
As a sentence in different languages are likely of varying lengths and word orders, the same semantic meaning will get encoded into different hidden state sequences.


There are two potential causes of this positional correspondence: residual connections and  encoder self-attention alignment.
We further hypothesize that, by modifying these two components accordingly, we can alleviate the positional correspondence.
Specifically, we set one encoder layer free from these constraints, so that it could create its own output ordering instead of always following a one-to-one mapping with its input.


\begin{figure}[t] 
\centering
    \centering
    \includegraphics[width=0.53\textwidth]{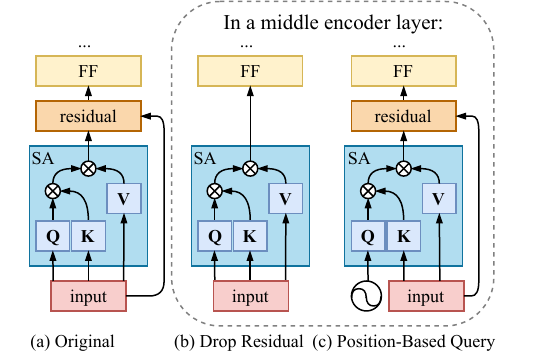}
\caption{\label{fig:3models} Illustrations of our proposed modifications to an original encoder layer: dropping residual connections once ($\S$\ref{subsec:residual}); making attention query based on position encoding ($\S$\ref{subsec:self_att}). Before each self-attention (SA) and feed forward (FF) layer we apply layer normalization, which is not visualized here for brevity.}
\end{figure}

\subsection{Modifying Residual Connections} \label{subsec:residual}
In the original Transformer architecture from \citet{transformer}, residual connections \cite{he2016deep} are applied in every layer, for both
the multihead attention 
and the feed-forward layer.
By adding the input embeddings to the layer outputs, the residual connections are devised to facilitate gradient flow to bottom layers of the network.
However, since the residual connections are present throughout all layers,
they strictly impose a one-to-one alignment between the inputs and outputs.
For the encoder, this causes the outputs to be positionally corresponding to the input tokens.

We propose to relax this condition, such that the encoder outputs becomes less position- and hence language-specific.
Meanwhile, to minimize the impact on the model architecture and ensure gradient flow, we limit this change to only one encoder layer, and only its multihead attention layer. Figure \ref{fig:3models}(b) gives a visualization of this change in comparison to the original encoder in Figure \ref{fig:3models}(a).

\subsection{Position-Based Self-Attention Query} \label{subsec:self_att}
Besides the residual connections, another potential reason for the positional correspondence is the encoder self-attention alignment.
Via the self-attention transform, each position is a weighted sum from all input positions.
While the weights theoretically can distribute over all input positions, they are often concentrated locally, particularly with output position $i$ focusing on input position $i$.
Previous works on various sequence tasks \cite{YANG2020121, zhang2020selfattention} have shown heavy weights on the diagonal of the encoder self-attention matrices.

In this light, the motivation of our method starts with the formation of the self-attention weight matrix: $\textrm{score}(\mathbf{Q}, \mathbf{K}) = \mathbf{Q} \mathbf{K} ^ {T}$,
where $\mathbf{Q}$ and $\mathbf{K}$ and the query and key matrices.
This $n \times n$ matrix encapsulates dot product at each position against all $n$ positions.
Since the dot product is used as a similarity measure, we hypothesize that when $\mathbf{Q}$ and $\mathbf{K}$ are similar, the matrix will have heavy weights on the diagonal, thereby causing the positional correspondence.
Indeed, $\mathbf{Q}$ and $\mathbf{K}$ are likely similar since they are projections from the same input.
We therefore propose to reduce this similarity by replacing the projection base of the self-attention query by a set of sinusoidal positional encodings. 
Moreover, to avoid possible interaction with positional information retained in $\mathbf{K}$,
we use a wave length for this set of sinusoidal encodings that is different from what is added onto encoder input embeddings. 
Figure \ref{fig:3models}(c) contrasts our position-based attention query with the original model in Figure \ref{fig:3models}(a), where the key, query, value are all projected from the input to the self-attention layer. 

\section{Experimental Setup}
Our experiments cover high- and low-resource languages and different data conditions.
We choose an English-centered setup, where we train on $\text{X} \leftrightarrow$ en parallel data, and test the zero-shot translation between all non-English languages.
This scenario is particularly difficult for zero-shot translation, as half of the  target-side training data is in English.
Indeed, recent works \cite{fan2020beyond, rios-etal-2020-subword} have outlined downsides of the English-centered configuration.
Nevertheless, intrigued by the potential of covering $N^2$ translation directions by training on $2N$ directions, we still explore this scenario.

\subsection{Datasets}
Our datasets originate from three sources: IWSLT 2017 \cite{cettolo2017overview}, Europarl v7 \cite{koehn2005europarl}, and PMIndia \cite{haddow2020pmindia}.
The IWSLT and Europarl data are taken from the MMCR4NLP corpus \cite{dabre2017mmcr4nlp}.
An overview of the datasets is in Table \ref{tab:dataset_overview}.

To investigate the role of training data diversity, we construct two conditions for Europarl, where 
one is fully multiway aligned, and the other has no multiway alignment at all.
Both are subsets of the full dataset with 1M parallel sentences per direction.
Moreover, we study the challenging case of PMIndia with little training data, distinct writing systems, and a large number of agglutinate languages that are specially difficult to translate into.
Table \ref{tab:language_overview} outlines the languages in our experiments.

\begin{table}[h]
\small
\centering
\begin{tabular}{ lccccc } 
\toprule
\textbf{Dataset}    & 
\textbf{X $\leftrightarrow$ en} & 
\shortstack[c]{\textbf{\# zero-shot} \\ \textbf{directions}}      & 
\shortstack[c]{\textbf{\# sent. per} \\ \textbf{direction}} \\ 
\midrule
\textbf{IWSLT}  & \{it, nl, ro\}
& 6                & 145K \\ 
\midrule
\multirow{3}{*}{\textbf{PMIndia}}                
& \multirow{3}{*}{\shortstack[l]{\{bn, gu, hi, \\ ml, mr, kn, \\or, te, pa\}}}          
& \multirow{3}{*}{72}               
& \multirow{3}{*}{26-53K}          
\\ 
\\
\\
\midrule
\textbf{Europarl}: \\
\hspace{4pt}w/o overlap & \multirow{3}{*}{\shortstack[l]{\{da, de, \\es, fi, fr, \\it, nl, pt\}}} &
\multirow{3}{*}{56} & 
119K \\
\hspace{4pt}multiway    &
&                  
& 
119K \\ 
\hspace{4pt}full    &
&
& 
1M \\
\bottomrule
\end{tabular}
\caption{\label{tab:dataset_overview} Overview of the datasets.}
\end{table}

\vspace*{-0.15in}

\begin{table}[h]
	\centering
	\small
	\setlength\tabcolsep{9pt} 
	\begin{tabular}{llllll}
		\toprule
		\textbf{Code} & \textbf{Name}   & \textbf{Family} & \textbf{Script} \\
		\midrule
        it      & Italian      & Romance & \multirow{9}{*}{Latin}\\
        nl      & Dutch        & Germanic\\
        ro      & Romanian     & Romance \\
        da      & Italian      & Romance & \\
        de      & German       & Germanic\\
        es      & Spanish      & Romance \\
        fi      & Finnish      & Uralic\\
        fr      & French       & Romance\\
        pt      & Portugese    & Romance\\
        \midrule
		bn      & Bengali       & Indo-Aryan    & Bengali       \\
		gu      & Gujarati      & Indo-Aryan    & Gujarati      \\
		hi      & Hindi         & Indo-Aryan    & Devanagari    \\
		kn      & Kannada       & Dravidian     & Kannada       \\
		ml      & Malayalam     & Dravidian     & Malayalam     \\
		mr      & Marathi       & Indo-Aryan    & Devanagari    \\
		or      & Odia          & Indo-Aryan    & Odia          \\
		pa      & Punjabi       & Indo-Aryan    & Gurmukhi      \\
		te      & Telugu        & Dravidian     & Telugu        \\
		\bottomrule
	\end{tabular}
	\caption{Overview of the languages in our experiments.}
	\label{tab:language_overview}
\end{table}

\subsection{Model Details and Baselines} \label{subsec:baselines}
\paragraph{Training Details}
By default we use Transformer \cite{transformer} with 5 encoder and decoder layers.
For the Europarl datasets with more training data, we enlarge the model to 8 encoder and decoder layers.
To control the output language, we use a target-language-specific begin-token as well as language embeddings concatenated with decoder word emebeddings\footnote{The concatenation of language embedding and decoder word embedding is then projected down to the embedding dimension to form the input embedding to the decoder.}, similar to \citet{pham-etal-2019-improving}.
We use 8 attention heads, embedding size of 512, inner size of 2048, dropout rate of 0.2, label smoothing rate of 0.1.
We use the learning rate schedule from \citet{transformer} with 8,000 warmup steps.
The source and target word embeddings are shared.
Furthermore, in the decoder, the parameters of the projection from hidden states to the vocabulary are tied with the transposition of the word lookup table.

Moreover, we include variational dropout \cite{varDrop} as a comparison since it was used in a previous work on zero-shot translation \cite{pham-etal-2019-improving} instead of the standard element-wise dropout.
With variational dropout, all timesteps in a layer output share the same mask. This differs from the standard dropout, where each element in each timestep is dropped according to the same dropout rate.
We hypothesize that this technique helps reduce the positional correspondence with input tokens by preventing the model from relying on specific word orders.

We train for 64 epochs and average the weights of the 5 best checkpoints ordered by dev loss. 
By default, we only include the supervised translation directions in the dev set. 
The only exception is the Europarl-full case, where we also include the zero-shot directions in dev set for early stopping.

When analyzing model hidden representations through classification performance (Subsection \ref{subsec:analyze_pos} and \ref{subsec:lan_independence}), we freeze the trained encoder-decoder weights and train the classifier for 5 epochs.
The classifier is a linear projection from the encoder hidden dimension to the number of classes, followed by softmax activation.
As the classification task is lightweight and convergence is fast, we reduce the warmup steps to 400 while keeping the learning rate schedule unchanged.

\paragraph{Proposed Models}
As motivated in Section \ref{sec:approach}, we modify the residual connections and the self-attention layer in a middle encoder layer.
Specifically, we choose the 3-rd and 5-th layer of the 5- and 8-layer models respectively.
We use ``Residual'' to indicate residual removal and ``Query'' the position-based attention query.
For the projection basis of the attention query, we use positional encoding with wave length 100.

\paragraph{Zero-Shot vs. Pivoting}
We compare the zero-shot translation performance with pivoting, i.e. directly translating the unseen direction $\text{X} \rightarrow \text{Y}$ vs. using English as an intermediate step, as in $\text{X} \rightarrow$ English $\rightarrow \text{Y}$.
The pivoting is done by the baseline multilingual model, which we expect to have similar performance to separately trained bilingual models.
For a fair comparison, in the Europarl-full case, pivoting is done by a baseline model trained till convergence with only supervised dev data rather than the early-stopped one.

\subsection{Preprocessing and Evaluation}
For the languages with Latin script, we first apply the Moses tokenizer and truecaser, and then learn byte pair encoding (BPE) using subword-nmt \cite{sennrich-etal-2016-neural}.
For the Indian languages, we use the IndicNLP library\footnote{\url{https://github.com/anoopkunchukuttan/indic_nlp_library}} and SentencePiece \cite{kudo-richardson-2018-sentencepiece} for tokenization and BPE respectively.
We choose 40K merge operations and only use tokens with minimum frequency of 50 in the training set.
For IWSLT, we use the official tst2017 set. 
For PMIndia, as the corpus does not come with dev and test sets, we partition the dataset ourselves by taking a multiway subset of all languages, resulting in 1,695 sentences in the dev and test set each.
For Europarl, we use the test sets in the MMCR4NLP corpus \cite{dabre2017mmcr4nlp}.
The outputs are evaluated by sacreBLEU\footnote{We use \texttt{BLEU+case.mixed+numrefs.1+smooth\\.exp+tok.13a+version.1.4.12} by default. On PMIndia, we use the SPM tokenizer (\texttt{tok.spm} instead of \texttt{tok.13a}) for better tokenization of the Indic languages. At the time of publication, the argument \texttt{tok.spm} is only available as a pull request to sacreBLEU: \url{https://github.com/mjpost/sacrebleu/pull/118}. We applied the pull request locally to use the SPM tokenizer.}  \cite{post-2018-call}.

\begin{table*}[h]
\small
\centering
\setlength\tabcolsep{7.4pt} 
\begin{tabular}{ llccccccrcccccc } 
\toprule
& \multirow{1}{*}{\shortstack[c]{ \\\\\\ \textbf{Dataset} }}  
          & \multicolumn{3}{c}{\textbf{Supervised Directions}} &&           \multicolumn{4}{c}{\textbf{Zero-Shot Directions}}\\
            \cmidrule{3-5}  \cmidrule{7-10}
&& Baseline & Residual & $+$Query && Pivot & Baseline & \multicolumn{1}{c}{Residual} & $+$Query\\
\midrule
$(1)$ &
 IWSLT
 & 29.8 & 29.4 & 29.4 & & 19.1 & 10.8 & 17.7 \hspace{2pt} ($+$6.9) & 17.8 \\
$(2)$ &
 Europarl multiway
 & 34.2 & 33.9 & 33.1 & & 25.9 & 11.3 & 26.1 ($+$14.8) & 25.1 \\
$(3)$ &
 Europarl w/o overlap
 & 35.6 & 35.4 & 34.9 & & 27.1 & \hspace{4pt}8.2 & 26.7 ($+$18.5) & 25.8 \\
$(4)$ &
 Europarl full
 & 35.4 & 36.4 & 35.9 & & 28.4 & 17.5 & 27.5 ($+$10.0) & 26.5 \\
$(5)$ &
 PMIndia
& 30.4 & 29.9 & 29.2 & & 22.1 & \hspace{4pt}0.8 & \hspace{2pt} 2.3 \hspace{2pt} ($+$1.5) &  \hspace{4pt} 1.1\\
\bottomrule
\end{tabular}
\caption{\label{tab:results_overview} BLEU\footnotemark scores on supervised and zero-shot directions. On IWSLT and Europarl (Row ($1$)-($4$)), removing residual connections once substantially improves zero-shot translation while retaining performance on supervised directions. On PMIndia (Row $5$), our approach can be improved further by additional regularization (Table \ref{tab:effect_of_vardrop}). }
\end{table*}
\begin{table}[t]
\small
\centering
\setlength\tabcolsep{4.5pt} 
\begin{tabular}{ llccc } 
\toprule
 \textbf{Dataset} & \textbf{Family} & \textbf{Baseline} & \textbf{Pivot} & \textbf{Residual} \\
\midrule
 \multirow{2}{*}{\shortstack[l]{Europarl \\multiway }} 
 & Germanic & \hspace{2pt} 6.9  & 25.9  & 26.2 ($+$0.3) \\
 & Romance  & 10.2 & 32.8  & 33.1 ($+$0.3) \\
 \midrule
 \multirow{2}{*}{\shortstack[l]{Europarl \\w/o overlap}}         
 & Germanic & 11.8  &   24.8 & 25.5 ($+$0.7) \\
 & Romance  & 13.5  &   31.0 & 32.3 ($+$1.3) \\
\bottomrule
\end{tabular}
\caption{\label{tab:results_similar_lan} Zero-shot BLEU scores between languages of the same families on Europarl multiway and non-overlap (Row ($2$) and ($3$) from Table \ref{tab:results_overview}). Our approach outperforms pivoting via English.}
\end{table}

\subsection{Adaptation Procedure}
To simulate the case of later adding a new language, we learn a new BPE model for the new language and keep the previous model unchanged.
Due to the increased number of unique tokens, the vocabulary of the previously-trained model is expanded.
In this case, for the model weights related to the word lookup table size, we initialize them as the average of existing embedding perturbed by random noise.

\section {Results} \label{sec:results}

\footnotetext[5]{Due to the large number of languages, we report the BLEU scores averaged over all directions here, and refer the readers to the appendix for detailed results.}

Our approach substantially improves zero-shot translation quality, as summarized in Table \ref{tab:results_overview}.
The first observation is that modification in residual connections is essential for zero-shot performance\footnote{We also experimented with: 1) removing the residual in more layers, but observed large negative impact on convergence; 2) replacing the residual connections by meanpooled sentence embeddings, but the gains on zero-shot directions were less than removing the residual connections.}.
We gain 6.9 and up to 18.5 BLEU points over the baseline on IWSLT and Europarl (Row $1$ to $4$) respectively.
When inspecting the model outputs, we see that the baseline often generates off-target translation in English, in line with observations from prior works \cite{arivazhagan2019missing, zhang-etal-2020-improving}. 
Our proposed models are not only consistent in generating the required target languages in zero-shot conditions, but also show competitive performance to pivoting via English.
The effects are particularly prominent between related languages. 
As shown in Table \ref{tab:results_similar_lan}, on Europarl, zero-shot outperforms the pivoting when translating between languages from the same families.
This is an attractive property especially when the computation resource is limited at inference time.

In the very challenging case of PMIndia (Row $5$), while removing residual does improve the zero-shot performance, the score of 2.3 indicates that the outputs are still far from being useful.
Nonetheless, we are able to remedy this by further regularization as we will present in Subsection \ref{subsec:additional_regularization}.

Contrary to the large gains by removing residual connections, the attention query modification is not effective when combined with residual removal.
This suggests that the primary source of position-specific representation is the residual connections.

Moreover, by contrasting Row $2$ and $3$ of Table \ref{tab:results_overview}, we show the effect of training data diversity.
In real-life, the parallel data from different language pairs are often to some degree multiway.
Multiway data could provide an implicit bridging that facilitates zero-shot translation.
With non-overlapping data, gains can come from training with a larger variety of sentences.
Given these two opposing hypotheses, our results suggest that the diverse training data is more important for both supervised and zero-shot performance.
With non-overlapping data, we first obverse improved supervised translation performance by around 1.5 points for all three model configurations (Baseline, Residual, Residual+Query).
Meanwhile, the zero-shot score also increases from 26.1 to 26.7 points with our model (Residual).
The baseline, on the contrary, loses from 11.3 to 8.2 points.
This suggests that our model can better utilize the diverse training data than the baseline under zero-shot conditions.

\subsection{Effect of Additional Regularization} \label{subsec:additional_regularization}
In Subsection \ref{subsec:baselines}, we hypothesized that variational dropout helps reduce position-specific representation.
Table \ref{tab:effect_of_vardrop} shows the outcome of replacing the standard dropout by this technique.
First, variational dropout also improves zero-shot performance over the baseline, yet not as strongly as residual removal.
On IWSLT and Europarl, there is no additive gain by combining both techniques.
On PMIndia, however, combining our model and variational dropout is essential for achieving reasonable zero-shot performance, as shown by the increase from 2.4 to 14.3 points.
Why is the picture different on PMIndia? 
We identify two potential reasons: 1) the low lexical overlap\footnote{We also tried mapping the 9 Indian languages into the Devanagari script, but got worse zero-shot performance compared to the current setup.} among the languages (8 different scripts in the 9 Indian languages);
2) the extreme low-resource condition (30K sentences per translation direction on average).

To understand this phenomenon, we create an artificial setup based on IWSLT with 1) no lexical overlap by appending a language tag before each token; 2) extremely low resource by taking a subset of 30K sentences per translation direction.
The scores in Table \ref{tab:iwslt_pmindia} show the increasing benefit of variational dropout given very low amount of training data and shared lexicon.
We interpret this through the lens of generalizable representations:
With low data amount or lexical overlap, the model tends to represent its input in a highly language-specific way, hence hurting zero-shot performance.

\begin{table}[h]
\small
\centering
\setlength\tabcolsep{5.5pt} 
\begin{tabular}{ lcc|ccccc|cc } 
\toprule
\multirow{1}{*}{\shortstack[c]{ \\\\\\ \textbf{Dataset} }}  
& \multicolumn{4}{c}{\textbf{Zero-Shot Directions}} \\ \cmidrule{2-5}
         & Baseline & $+$vardrop & Residual & $+$vardrop \\
\midrule
IWSLT    &    10.8 & 14.9 & \textbf{17.7} & 17.7 \\
Europarl &  \hspace{5pt}8.2 & 25.1 & \textbf{26.7} & 26.4 \\
PMIndia  &  \hspace{5pt}0.8 & \hspace{5pt}2.3  &  \hspace{5pt}2.4 & \textbf{14.3} \\
\bottomrule
\end{tabular}
\caption{\label{tab:effect_of_vardrop} Zero-shot BLEU scores by variational dropout (``$+$vardrop'') on IWSLT, Europarl non-overlap, and PMIndia.
On the first two datasets, combining residual removal and variational dropout has no synergy.
On PMIndia with little data and low lexical overlap, the combination of the two is essential.}
\end{table}

\begin{table}[h]
\small
\begin{tabular}{ llcccccc } 
\toprule
      & \textbf{Condition}         & \textbf{Residual}      & $+$ \textbf{vardrop} \\
\midrule
($1$) & Normal                      & 17.7                      & 17.7 ($+$0.0) \\
($2$) & $(1)+$little data           & 11.9                      & 12.9 ($+$\textbf{1.0}) \\
($3$) & $(2)+$no lexical overlap    & \hspace{4pt}9.7           & 12.2 ($+$\textbf{2.5}) \\
\bottomrule
\end{tabular}
\caption{\label{tab:iwslt_pmindia} 
Zero-shot BLEU scores of on a subset of IWSLT artificially constructed with little training data and no shared lexicon. 
The benefit of regularizing by variational dropout becomes prominent as the amount of training data and shared lexicon decreases.}
\end{table}


\subsection{Adaptation to Unseen Language}
So far our model has shown promising zero-shot performance.
Here we extend the challenge of zero-shot translation by integrating a new language.
Specifically, we finetune a trained English-centered many-to-many system with a new language using a small amount of $\text{X}_\textnormal{new}\leftrightarrow$ English parallel data. 
At test time, we perform zero-shot translation between $\text{X}_\textnormal{new}$ and all non-English languages previously involved in training.
This practically simulates the scenario of later acquiring parallel data between a low-resource language and the central bridging language in an existing system.
After finetuning with the new data, we can potentially increase translation coverage by $2N$ directions, with $N$ being the number of languages originally in training.
We finetune a trained system on IWSLT (Row $1$ in Table \ref{tab:results_overview}) using a minimal amount of de $\leftrightarrow$ en data with 14K sentences.
When finetuning we include the original $\text{X}_\textnormal{old}\leftrightarrow$ en training data, as otherwise the model would heavily overfit. 
This procedure is relatively lightweight, since the model has already converged on the original training data. 

In Table \ref{tab:few_shot}, our model outperforms the baseline on zero-shot translation, especially when translating from the new language ($\text{X}_{\textnormal{new}}$ $\rightarrow$).
When inspecting the outputs, we see the baseline almost always translate into the wrong language (English), causing the low score of 1.8.
We hypothesize that the baseline overfits more on the supervised direction ($\text{X}_{\textnormal{new}}$ $\rightarrow$ en), where it achieves the higher score of 18.5.
In contrast, our model is less susceptible to this issue and consistently stronger under zero-shot conditions.

\begin{table}[h]
\small
\setlength\tabcolsep{5.5pt} 
\begin{tabular}{ lccccc } 
\toprule
&\multicolumn{2}{c}{\textbf{Supervised}} && \multicolumn{2}{c}{\textbf{Zero-Shot}}\\
\cmidrule{2-3}  \cmidrule{5-6}
&Baseline & Residual && Baseline & Residual \\
\midrule
 $\text{X}_{\text{new}}$ $\rightarrow$ &18.5     & 17.3     && 1.8 & \hspace{4pt}6.7 \\
$\rightarrow$ $\text{X}_{\text{new}}$  &13.6     & 13.4     && 8.3 & 10.7 \\
\bottomrule
\end{tabular}
\caption{\label{tab:few_shot} Effects of adaptation to new language (de $\leftrightarrow$ en  on IWSLT. Zero-shot translation directions are de $\leftrightarrow$ \{it, nl, ro\}. Our model has significantly stronger zero-shot performance.}
\end{table}

\section{Discussions and Analyses}
To see beyond BLEU scores, we first analyze how much position- and language-specific information is retained in the encoder hidden representations before and after applying our approaches.
We then study circumstances where zero-shot translation tends to outperform its pivoting-based counterpart.
Lastly, we discuss the robustness of our approach to the impact of different implementation choices.

\subsection{Inspecting Positional Correspondence} \label{subsec:analyze_pos}
To validate whether the improvements in zero-shot performance indeed stem from less positional correspondence to input tokens,
we assess the difficulty of recovering input positional information before and after applying our proposed method.
Specifically, we train a classifier to predict the input token ID's (which word it is) or position ID's (the word's absolute position in a sentence) based on encoder outputs.
Such prediction tasks have been used to analyze linguistic properties of encoded representation \cite{adi2017fine}.
Our classifier operates on each timestep and uses a linear projection from the embedding dimension to the number of classes, i.e. number of unique tokens in the vocabulary or number of maximum timesteps.

Table \ref{tab:position_classification} compares the classification accuracy of the baseline and our model.
First, the baseline encoder output has an exact one-to-one correspondence to the input tokens, as evidenced by the nearly perfect accuracy when recovering token ID's.
This task becomes much more difficult under our model.
We see a similar picture when recovering the position ID's.

\begin{table}[h]
\small
\centering
\setlength\tabcolsep{8pt} 
\begin{tabular}{ lccc } 
\toprule
\textbf{Dataset} & 
\textbf{Model} & 
\textbf{Token ID} & 
\textbf{Position ID} \\
\midrule
IWSLT       & Baseline & 99.9\%     & 93.3\% \\
            & Residual & 48.5\%     & 51.4\% \\
\midrule
Europarl    & Baseline & 99.5\%     & 85.1\% \\
non-overlap & Residual & 71.6\%     & 22.5\% \\
\midrule
PMIndia     & Baseline & 99.6\%     & 90.1\% \\
            & Residual & 63.3\%     & 26.9\% \\
\bottomrule
\end{tabular}
\caption{\label{tab:position_classification} 
Accuracy of classifiers trained to recover input positional information (\textbf{token} ID or \textbf{position} ID) based on encoder outputs.
Lower values indicate higher difficulty of recovering the information, and therefore less positional correspondence to the input tokens.}
\end{table}


\begin{figure}[h]
	\pgfplotsset{compat=1.11,
		/pgfplots/ybar legend/.style={
			/pgfplots/legend image code/.code={%
				\draw[##1,/tikz/.cd,yshift=-0.25em]
				(0cm,0cm) rectangle (3pt,0.8em);},
		},
	}
	
	\begin{tikzpicture}
	\begin{axis}[
	axis x line*=bottom,
	axis y line*=left,
	width  = 0.45\textwidth,
	height = 2.7cm,
	major x tick style = transparent,
	ybar=0pt,
	bar width=10pt,
	symbolic x coords={layer 1, layer 2, layer 3, layer 4, layer 5},
	xtick = data,
	enlarge x limits=0.2,
	ymax = 100,
	ymin = 0,
	ytick={0, 25, 50, 75, 100},
	ymajorgrids = true,
	ylabel style = {align=center},
	ylabel = Accuracy \\ ($\%$),
	ylabel shift=-6pt,
	legend columns=-1,
	legend style={
		at={(0.5,1.5)},
		anchor=north,
		column sep=1ex,
		draw=none
	}
	]
	\small
	\addplot[style={fill=darkgray,mark=none,draw=none,draw opacity=0}]
	coordinates {(layer 1, 99.8) (layer 2, 99.6108829975128) (layer 3, 98.2299149036408) (layer 4, 95.5880999565124) (layer 5, 92.6758408546448)};
	
	\addplot[style={fill=LightGray,mark=none,draw=none,draw opacity=0}]
	coordinates {(layer 1, 99.9) (layer 2, 99.4202792644501) (layer 3, 52.9687225818634) (layer 4, 50.2031862735748) (layer 5, 44.2916423082352)};
	
	\legend{Baseline, Residual}
	\end{axis}
	\end{tikzpicture}
	\caption{\label{fig:accuracy_per_layer} Accuracy of recovering position ID's after each encoder layer on IWSLT. When we remove the residual connection in the 3rd encoder layer, classification is much more difficult.}
\end{figure}
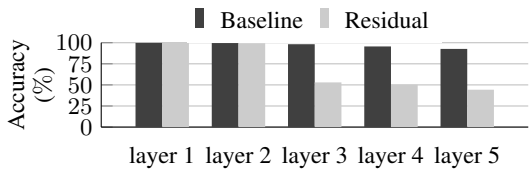

We also try to recover the position ID's based on the outputs from each layer.
As shown in Figure \ref{fig:accuracy_per_layer}, the accuracy drops sharply at the third layer, where the residual connection is removed.
This shows that the devised transition point at a middle encoder layer is effective.

\subsection{Inspecting Language Independence} \label{subsec:lan_independence}
To test whether our model leads to more language-independent representations, we assess the similarity of encoder outputs on the sentence and token level using the two following methods:

\paragraph{SVCCA} 
The singular vector canonical correlation analysis (SVCCA; \citeauthor{SVCCA}, \citeyear{SVCCA}) measures similarity of neural network outputs, and has been used to assess representational similarity in NMT \cite{kudugunta-etal-2019-investigating}.
As SVCCA operates on fixed-size inputs, we meanpool the encoder outputs and measure similarity on a sentence level.


\paragraph{Language Classification Accuracy}
Since more similar representations are more difficult to distinguish, poor performance of a language classifier indicates high similarity.
Based on a trained model, we learn a token-level linear projection from the encoder outputs to the number of classes (languages).

\begin{figure}[h]
\centering
\includegraphics[width=0.42\textwidth]{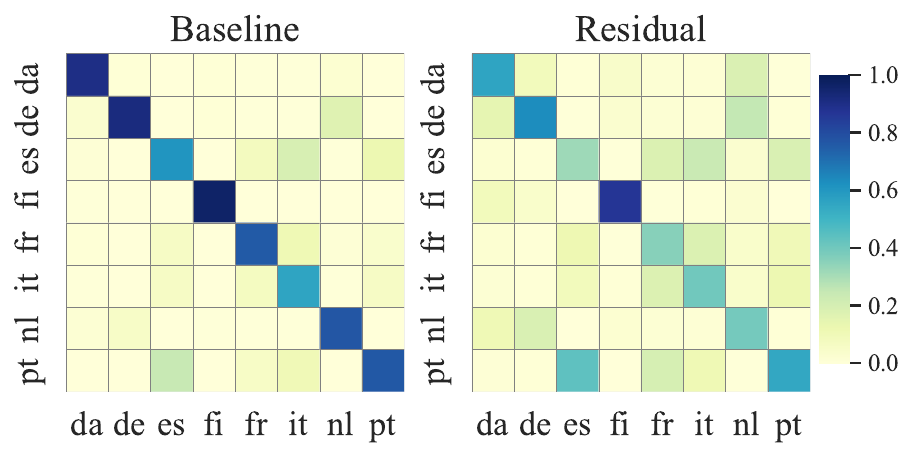}
\caption{\label{fig:confusion} Confusion matrices when classifying languages in Europarl non-overlap (x: true, y: predicted). Encoder outputs of related languages (Romance / Germanic) are more similar after residual removal.}
\end{figure}

\paragraph{Findings}
As shown in Table \ref{tab:similarity}, our model consistently achieves higher SVCCA scores and lower classification accuracy than the baseline, indicating more language-independent representations.
When zooming into the difficulty of classifying the languages, we further notice much higher confusion (therefore similarity) between related languages.
For instance, Figure \ref{fig:confusion} shows the confusion matrix when classifying the 8 source languages in Europarl.
After residual removal, the similarity is much higher within the Germanic and Romance family.
This also corresponds to cases where our model outperforms pivoting (Table \ref{tab:results_similar_lan}).

\begin{table}[h]
\small
\centering
\setlength\tabcolsep{6.5pt} 
\begin{tabular}{ lcccc } 
\toprule
\shortstack[c]{\textbf{Dataset} \\ {\color{white}$\uparrow$} }     & 
\shortstack[c]{\textbf{Model} \\ {\color{white}$\uparrow$} }   & 
\shortstack[c]{\textbf{SVCCA Score} \\ $\uparrow$}    &
\shortstack[c]{\textbf{Accuracy} \\ $\downarrow$} \\
\midrule
IWSLT       & Baseline & 0.682            &  95.9\%   \\   
            & Residual & 0.703            &  87.6\%   \\
\midrule
Europarl    & Baseline & 0.652            &  87.0\%   \\
non-overlap & Residual & 0.680            &  69.9\%   \\
\midrule
PMIndia     & Baseline & 0.621            &  74.1\%   \\
            & Residual & 0.650            &  62.0\%   \\
\bottomrule
\end{tabular}
\caption{\label{tab:similarity} 
Average pairwise similarity of encoder outputs for between all languages in each dataset.
Higher SVCCA scores and lower classification accuracy indicate higher similarity. We note that the SVCCA score between random vectors is around 0.57.}
\end{table}

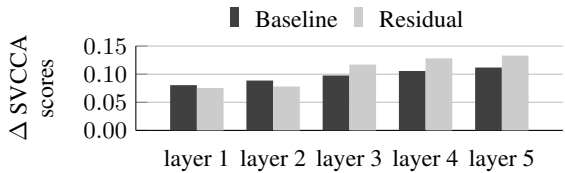
\begin{figure}[h]
	\pgfplotsset{compat=1.11,
		/pgfplots/ybar legend/.style={
			/pgfplots/legend image code/.code={%
				\draw[##1,/tikz/.cd,yshift=-0.25em]
				(0cm,0cm) rectangle (3pt,0.8em);},
		},
	}
	
	\begin{tikzpicture}
	\begin{axis}[
	axis x line*=bottom,
	axis y line*=left,
	width  = 0.45\textwidth,
	height = 2.7cm,
	major x tick style = transparent,
	ybar=0pt,
	bar width=10pt,
	symbolic x coords={layer 1, layer 2, layer 3, layer 4, layer 5},
	xtick = data,
	enlarge x limits=0.2,
	ymax = 0.15,
	ymin = 0,
	scaled y ticks=false,
	yticklabel=\pgfkeys{/pgf/number format/.cd,fixed,precision=2,zerofill}\pgfmathprintnumber{\tick},
	ytick={0, 0.05, 0.10, 0.15},
	ymajorgrids = true,
	ylabel style = {align=center},
	ylabel = $\Delta$ SVCCA \\ scores,
	ylabel shift=5pt,
	legend columns=-1,
	legend style={
		at={(0.5,1.55)},
		anchor=north,
		column sep=1ex,
		draw=none
	}
	]
	\small
	\addplot[style={fill=darkgray,mark=none,draw=none,draw opacity=0}]
	coordinates {(layer 1, 0.08046666667) (layer 2, 0.08863333333) (layer 3, 0.09766666667) (layer 4, 	0.1056833333) (layer 5, 0.1118)};
	
	\addplot[style={fill=LightGray,mark=none,draw=none,draw opacity=0}]
	coordinates {(layer 1, 0.07545) (layer 2, 0.07803333333) (layer 3, 0.11705) (layer 4, 		0.1281) (layer 5, 0.13305)};
	
	\legend{Baseline, Residual}
	\end{axis}
	\end{tikzpicture}
\caption{\label{fig:svcca_per_layer} Average SVCCA scores after each encoder layer between all language pairs in IWSLT (\{it, nl, ro, en\}).
Scores are reported additive to scores between random vectors.
Similarity significantly increases after the 3rd layer where we apply residual removal.}
\end{figure}

Moreover, we compare the SVCCA scores after each encoder layer, as shown in Figure \ref{fig:svcca_per_layer}.
Confirming our hypotheses, the model outputs are much more similar after the transition layer, as shown by the sharp increase at layer 3.
This contrasts the baseline, where similarity increases nearly linearly.

\vspace*{-0.1in}

Given these findings and previous analyses in Subsection \ref{subsec:analyze_pos}, we conclude that our devised changes in a middle encoder layer allows higher cross-lingual generalizability in top layers while retaining the language-specific bottom layers.

\subsection{Understanding Gains of Zero-Shot Translation Between Related Languages}
In Subsection \ref{sec:results} we have shown that between related languages zero-shot translation surpasses pivoting performance.
Here we manually inspect some pivoting translation outputs (nl$\rightarrow$en$\rightarrow$de) and compare them to zero-shot outputs (de$\rightarrow$en).
In general, we observe that the translations without pivoting are much more similar to the original sentences.
For instance in Table \ref{tab:results_similar_lan}, when pivoting, the Dutch sentence ``\textit{geven het voorbeeld} (give the example)'' is first translated to ``set the example'', then to ``\textit{setzen das Beispiel} (set the example)'' in German, which is incorrect as the verb ``\textit{setzen} (set)'' cannot go together with the noun ``\textit{Beispiel} (example)''.
The zero-shot outputs, on the other hand, directly translates ``\textit{geven} (give; Dutch)'' to ``\textit{geben} (give; German)'', resulting in a more natural pairing with ``\textit{Beispiel} (example)''.
With this example, we intend to showcase the potential of bypassing the pivoting step and better exploiting language similarity.

\begin{table}[h]
\small
\centering
\setlength\tabcolsep{8pt} 
\begin{tabular}{ l|lrrrrrrrrrrrr } 
\toprule
\shortstack[l]{Input \\(nl)}                    
& \shortstack[l]{... geven in dit verband het verkeerde \\ voorbeeld, maar anderen helaas ook.}   \\
\arrayrulecolor{lightgray} 
\midrule
\shortstack[l]{Pivot-in \\(nl$\rightarrow$en)}  
& \shortstack[l]{... are setting the wrong example here, \\ but others are unfortunately also.} \\ 
\midrule
\shortstack[l]{Pivot-out \\(en$\rightarrow$de)} 
& \shortstack[l]{... \textbf{setzen} hier das falsche Beispiel ein, \\ andere sind leider auch.}\\
 \arrayrulecolor{black} 
\midrule
\shortstack[l]{Zero-shot \\(nl$\rightarrow$de)} 
& \shortstack[l]{... geben in diesem Zusammenhang das \\ falsche Beispiel, aber leider auch andere.}\\
\bottomrule
\end{tabular}
\caption{\label{tab:example_pivoting} An example of pivoting (nl$\rightarrow$en$\rightarrow$de) vs zero-shot (nl$\rightarrow$de).
Pivoting via English leads to the incorrect verb-noun pairing of ``\textit{setzen das Beispiel} (set the example)'' in German, while zero-shot output utilizes language similarity to get higher output quality.}
\end{table}

\vspace*{-0.1in}
\subsection{Where to Remove Residual Connections}
In our main experiments, all proposed modifications take place in a middle encoder layer.
After comparing the effects of residual removal in each of the encoder layers, our first observation is that the bottom encoder layer should remain fully position-aware.
Removing the residual connections in the first encoder layer degrades zero-shot performance by 2.8 BLEU on average on IWSLT.
Secondly, leaving out residual connections in top encoder layers (fourth or fifth layer of the five layers) slows down convergence. When keeping the number of training epochs unchanged from our main experiments, it comes with a loss of 0.4 BLEU on the supervised directions.
This is likely due to the weaker gradient flow to the bottom layers.
The two observations together support our choice of using the middle encoder layer as a transition point.


\subsection{Learned or Fixed Positional Embedding}
While we use fixed trigonometric positional encodings in our main experiments, we also validate our findings with learned positional embeddings on the IWSLT dataset. 
First, the baseline still suffers from off-target zero-shot translation (average BLEU scores on supervised directions: 29.6; zero-shot: 4.8). Second, removing the residual connection in a middle layer is also effective in this case (supervised: 29.1; zero-shot: 17.1).
These findings suggest that our approach is robust to the form of positional embedding.
Although learned positional embeddings are likely more language-agnostic by seeing more languages, as we still present source sentences as a sequence of tokens, the residual connections, when present in all layers, would still enforce a one-to-one mapping to the input tokens.
This condition allows our motivation and approach to remain applicable.

\vspace*{-0.04in}
\section{Related Work}
\vspace*{-0.04in}
Initial works on multilingual translation systems already showed some zero-shot capability \cite{johnson-etal-2017-googles, ha2016toward}.
Since then, several works improved zero-shot translation performance by controlling or learning the level of parameter sharing between languages \cite{lu-etal-2018-neural, platanios-etal-2018-contextual}.

Recently, models with full parameter sharing have gained popularity, with massively multilingual systems showing encouraging results \cite{aharoni-etal-2019-massively, zhang-etal-2020-improving, fan2020beyond}.
Besides advantages such as compactness and ease of deployment, the tightly-coupled model components also open up new questions.
One question is how to form language-agnostic representations at a suitable abstraction level.
In this context, one approach is to introduce auxiliary training objectives to encourage similarity between the representations of different languages \cite{arivazhagan2019missing, pham-etal-2019-improving}.
In this work we took a different perspective:
Instead of introducing additional objectives, we relax some of the pre-defined structure to facilitate language-independent representations.

Another line of work on improving zero-shot translation utilizes monolingual pretraining \cite{gu-etal-2019-improved, ji2020cross} or synthetic data for the zero-shot directions by generated by backtranslation \cite{gu-etal-2019-improved, zhang-etal-2020-improving}.
With both approaches, the zero-shot directions must be known upfront in order to train on the corresponding languages.
In comparison, our adaptation procedure offers more flexibility, as the first training step remains unchanged regardless of which new language is later finetuned on.
This could be suitable to the practical scenario of later acquiring data for the new language.
Our work is also related to adaptation to new languages.
While the existing literature mostly focused on adapting to one or multiple supervised training directions \cite{zoph-etal-2016-transfer, neubig-hu-2018-rapid, zhou-etal-2019-handling, murthy-etal-2019-addressing, bapna-firat-2019-simple}, our focus in this work is to rapidly expand translation coverage via zero-shot translation.

While our work concentrates on an English-centered data scenario, another promising direction to combat zero-shot conditions is to enrich available training data by mining parallel data between non-English languages \cite{fan2020beyond, freitag-firat-2020-complete}.
On a broader scope of sequence-to-sequence tasks, \citet{dalmia2019enforcing} enforced encoder-decoder modularity for speech recognition.
The goal of modular encoders and decoders is analogous to our motivation for zero-shot translation.

\vspace*{-0.03in}
\section {Conclusion}
\vspace*{-0.03in}
In this work, we show that 
the positional correspondence to input tokens hinders zero-shot translation.
Specifically, we demonstrate that:
1) the encoder outputs retain word orders of source languages;
2) this positional information reduces cross-lingual generalizability and therefore zero-shot translation quality;
3) the problems above can be easily alleviated by removing the residual connections in one middle encoder layer. 
With this simple modification, we achieve improvements up to 18.5 BLEU points on zero-shot translation.
The gain is especially prominent in related languages, where our proposed model outperforms pivot-based translation.
Our approach also enables integration of new languages with little parallel data. 
Similar to interlingua-based models, by adding two translation directions, we can increase the translation coverage by 2$N$ language pairs, where $N$ is the original number of languages.
In terms of model representation, we show that the encoder outputs under our proposed model are more language-independent both on a sentence and token level.

\vspace*{-0.05in}
\section*{Acknowledgment}
\vspace*{-0.05in}
This work is supported by a Facebook Sponsored Research Agreement. We thank Yuqing Tang for helpful comments, and Ngoc Quan Pham for sharing the training details of \cite{pham-etal-2019-improving}.

\vspace*{-0.05in}
\section*{Broader Impact}
\vspace*{-0.05in}
We proposed approaches to improve zero-shot translation, which is especially suitable to low-resource scenarios with no training data available between some languages. We also validated our approaches on actual low-resource languages. However, as the models are trained on single domains, when facing out-of-domain test sentences, they could suffer from hallucination, i.e. produce translations unrelated to the input sentences.

\bibliography{anthology,acl2020}
\bibliographystyle{acl_natbib}

\appendix
\onecolumn 
\renewcommand{\arraystretch}{0.915}
\section{Appendix: BLEU Scores per Translation Direction}
\subsection{Multiway IWSLT 2017}
\begin{center}
\scriptsize
\begin{tabular}{ llrrrrrrrrr} 
\toprule
& Direction & Pivot (X$\rightarrow$en$\rightarrow$Y) & Baseline & $+$vardrop & Residual & $+$vardrop & Residual$+$Query & $+$vardrop \\ 
\midrule 
\multirow{7}{*}{\rotatebox[origin=c]{90}{\bf{Supervised}}} 
& en-it             & - & 30.4 &   30.1 &   29.8 &   29.8 &   30.0 &   29.7 \\ 
& en-nl             & - & 27.9 &   27.7 &   27.4 &   27.6 &   27.4 &   26.9 \\
& en-ro             & - & 23.2 &   22.9 &   22.9 &   22.4 &   23.3 &   22.2 \\
& it-en             & - & 35.8 &   35.7 &   35.5 &   35.0 &   35.2 &   33.8 \\ 
& nl-en             & - & 31.0 &   31.0 &   30.4 &   30.5 &   30.5 &   29.9 \\
& ro-en             & - & 30.6 &   30.4 &   30.1 &   29.4 &   29.8 &   29.3 \\
& \multicolumn{2}{l}{\textbf{Average}}
                        &  \textbf{29.8} & \textbf{29.6} & \textbf{29.4} & \textbf{29.1} & \textbf{29.4} & \textbf{28.6} \\
\midrule
\multirow{7}{*}{\rotatebox[origin=c]{90}{\bf{Zero-shot}}} 
& it-nl             & 19.8 & 11.5 &   15.0 &   18.5 &   18.6 &   18.7 & 18.0 \\
& it-ro             & 18.3 & 11.8 &   15.8 &   17.8 &   17.7 &   18.2 & 17.3 \\
& nl-it             & 20.1 & 10.0 &   15.2 &   17.9 &   18.1 &   17.9 & 18.0 \\
& nl-ro             & 16.4 &  9.2 &   14.1 &   15.5 &   15.5 &   15.6 & 15.5 \\
& ro-it             & 21.1 & 12.2 &   15.8 &   19.6 &   19.8 &   19.5 & 19.6 \\
& ro-nl             & 18.5 & 10.3 &   13.4 &   16.8 &   16.6 &   16.7 & 16.5 \\
&\textbf{Average}
                    & \textbf{19.1} & \textbf{10.8} & \textbf{14.9} & \textbf{17.7} & \textbf{17.7} & \textbf{17.8} & \textbf{17.5} \\
\bottomrule
\end{tabular}
\end{center}


\subsection{Europarl}
\subsubsection{Non-Overlapping Data}
{
\scriptsize
\begin{longtable}{ llrrrrrrrr }
\toprule
& Direction & Pivot (X$\rightarrow$en$\rightarrow$Y) & Baseline & $+$vardrop & Residual & $+$vardrop & Residual$+$Query & $+$vardrop \\
\midrule
\endhead
\multirow{17}{*}{\rotatebox[origin=c]{90}{\bf{Supervised}}} 
& da-en & - & 38.3 & 38.5 & 38.1 & 38.2 & 37.7 & 37.3 \\
& de-en & - & 36.1 & 36.0 & 35.9 & 35.5 & 35.2 & 34.9 \\
& es-en & - & 43.0 & 43.2 & 43.0 & 42.9 & 42.5 & 42.1 \\
& fi-en & - & 32.6 & 32.8 & 32.4 & 31.9 & 31.9 & 31.0 \\
& fr-en & - & 39.4 & 39.5 & 39.1 & 39.0 & 38.8 & 37.5 \\
& it-en & - & 37.3 & 36.9 & 36.7 & 36.4 & 36.4 & 35.2 \\
& nl-en & - & 34.4 & 34.2 & 34.0 & 33.8 & 33.3 & 32.9 \\
& pt-en & - & 41.1 & 41.0 & 40.7 & 40.6 & 40.2 & 39.5 \\
& en-da & - & 36.5 & 36.7 & 36.2 & 36.1 & 35.8 & 35.7 \\
& en-de & - & 28.1 & 28.0 & 27.7 & 27.5 & 27.5 & 26.7 \\
& en-es & - & 42.5 & 42.6 & 42.2 & 42.2 & 41.7 & 41.6 \\
& en-fi & - & 22.5 & 22.2 & 22.2 & 21.6 & 21.3 & 20.8 \\
& en-fr & - & 37.7 & 38.0 & 37.8 & 37.5 & 37.4 & 37.0 \\
& en-it & - & 32.7 & 32.7 & 32.4 & 32.2 & 32.0 & 31.8 \\
& en-nl & - & 29.7 & 29.9 & 29.6 & 29.4 & 29.5 & 29.1 \\
& en-pt & - & 38.3 & 38.4 & 38.0 & 38.0 & 37.9 & 37.6\\
& \textbf{Average} &  &  \textbf{35.6} & \textbf{35.7} & \textbf{35.4} & \textbf{35.2} & \textbf{34.9} & \textbf{34.4}\\
\midrule
\multirow{57}{*}{\rotatebox[origin=c]{90}{\bf{Zero-shot}}} 
& da-de & 24.2 &  5.2 & 21.9 & 25.0 & 24.5 & 23.9 & 23.7 \\* 
& da-es & 33.1 & 15.8 & 32.4 & 32.8 & 32.5 & 32.3 & 31.6 \\*
& da-fi & 18.1 &  5.0 & 16.6 & 18.1 & 17.1 & 17.2 & 16.6 \\*
& da-fr & 30.6 &  9.9 & 28.1 & 29.4 & 29.3 & 28.9 & 28.4 \\*
& da-it & 26.1 & 11.5 & 24.3 & 24.8 & 25.0 & 24.8 & 24.3 \\*
& da-nl & 26.3 &  6.5 & 24.3 & 25.8 & 26.0 & 25.5 & 25.5 \\*
& da-pt & 29.9 & 14.4 & 28.6 & 29.3 & 29.1 & 28.8 & 28.4 \\*

& de-da & 29.2 &  9.5 & 26.9 & 29.2 & 29.1 & 27.5 & 28.3 \\*
& de-es & 32.1 & 11.6 & 31.2 & 32.1 & 31.6 & 31.0 & 30.9 \\*
& de-fi & 17.9 &  4.2 & 16.7 & 17.5 & 16.7 & 16.3 & 16.1 \\*
& de-fr & 29.9 &  7.3 & 27.8 & 28.9 & 29.2 & 28.4 & 28.6 \\*
& de-it & 25.6 &  8.8 & 23.4 & 24.6 & 24.5 & 23.7 & 23.6 \\*
& de-nl & 25.9 &  5.5 & 23.7 & 26.1 & 26.1 & 25.2 & 25.4 \\*
& de-pt & 29.2 & 10.8 & 28.0 & 28.9 & 28.4 & 28.2 & 27.9 \\*

& es-da & 31.4 & 10.2 & 28.1 & 31.0 & 30.5 & 29.5 & 29.6 \\
& es-de & 24.8 &  4.3 & 21.4 & 24.3 & 23.8 & 23.4 & 23.1 \\
& es-fi & 19.4 &  4.5 & 17.3 & 18.6 & 17.5 & 17.7 & 17.0 \\
& es-fr & 34.8 &  9.1 & 33.4 & 34.8 & 34.9 & 34.2 & 34.1 \\ 
& es-it & 29.6 &  9.8 & 28.7 & 29.7 & 29.6 & 29.1 & 28.9 \\
& es-nl & 26.8 &  5.9 & 24.3 & 26.5 & 26.1 & 25.3 & 25.5 \\
& es-pt & 35.1 & 13.6 & 35.3 & 35.9 & 36.0 & 35.5 & 35.5 \\ 

& fi-da & 26.3 &  9.8 & 23.6 & 25.4 & 24.9 & 24.1 & 24.3 \\
& fi-de & 20.8 &  4.0 & 17.8 & 20.2 & 19.4 & 18.8 & 18.9 \\
& fi-es & 29.2 & 11.9 & 27.8 & 28.4 & 27.6 & 27.3 & 26.6 \\
& fi-fr & 27.2 &  7.6 & 24.3 & 25.4 & 25.4 & 24.7 & 24.3 \\
& fi-it & 23.2 &  8.4 & 20.9 & 21.8 & 21.4 & 20.9 & 20.1 \\
& fi-nl & 22.8 &  5.3 & 20.2 & 22.0 & 21.6 & 20.7 & 20.9 \\
& fi-pt & 26.5 & 10.8 & 25.0 & 25.3 & 24.8 & 24.0 & 23.9 \\

& fr-da & 29.4 &  8.6 & 25.6 & 28.6 & 28.3 & 26.8 & 26.4 \\
& fr-de & 23.6 &  3.4 & 19.3 & 23.1 & 22.7 & 21.6 & 21.3 \\
& fr-es & 35.9 & 11.8 & 35.5 & 36.4 & 36.2 & 35.1 & 34.7 \\ 
& fr-fi & 17.9 &  3.9 & 15.7 & 17.2 & 16.0 & 15.8 & 15.2 \\
& fr-it & 28.4 &  8.0 & 27.3 & 28.7 & 28.3 & 27.7 & 26.8 \\ 
& fr-nl & 26.3 &  5.0 & 22.8 & 25.9 & 25.4 & 24.9 & 24.7 \\
& fr-pt & 32.8 & 11.5 & 31.8 & 33.3 & 33.1 & 32.6 & 31.9 \\ 

& it-da & 27.6 &  8.3 & 24.3 & 26.8 & 26.7 & 25.2 & 25.4 \\
& it-de & 22.3 &  3.5 & 18.1 & 21.3 & 21.1 & 20.2 & 20.0 \\
& it-es & 33.7 & 10.9 & 33.9 & 34.4  & 34.4 & 33.5 & 33.0 \\ 
& it-fi & 17.0 &  3.4 & 13.4 & 15.9 & 13.8 & 14.6 & 13.0 \\
& it-fr & 31.4 &  6.9 & 30.2 & 31.6 & 31.4 & 30.8 & 30.7 \\
& it-nl & 24.9 &  5.0 & 21.6 & 24.0 & 23.7 & 23.4 & 23.3 \\
& it-pt & 30.8 & 10.6 & 30.4 & 31.3 & 31.1 & 30.9 & 30.7 \\ 

& nl-da & 27.7 & 10.1 & 25.5 & 27.7 & 27.8 & 26.6 & 27.0 \\
& nl-de & 23.2 &  4.4 & 20.0 & 23.5 & 23.4 & 22.5 & 22.7 \\ 
& nl-es & 30.7 & 12.6 & 30.0 & 30.8 & 30.6 & 30.1 & 29.8 \\
& nl-fi & 16.5 &  4.7 & 15.5 & 16.0 & 15.8 & 15.6 & 15.4 \\
& nl-fr & 28.8 &  8.3 & 27.1 & 28.1 & 28.3 & 27.6 & 27.8 \\
& nl-it & 24.5 &  9.2 & 22.5 & 23.9 & 23.8 & 23.1 & 23.3 \\
& nl-pt & 28.2 & 12.2 & 27.0 & 27.5 & 27.3 & 27.2 & 27.1 \\

& pt-da & 30.5 &  9.5 & 27.1 & 29.5 & 29.2 & 28.0 & 28.4 \\
& pt-de & 24.3 &  3.5 & 20.6 & 23.6 & 23.2 & 22.4 & 22.5 \\
& pt-es & 37.7 & 14.3 & 37.7 & 38.3 & 38.2 & 37.5 & 37.4 \\ 
& pt-fi & 18.4 &  4.1 & 16.8 & 17.9 & 17.1 & 17.2 & 16.6 \\
& pt-fr & 34.1 &  7.7 & 32.6 & 33.9 & 34.1 & 33.4 & 33.3 \\
& pt-it & 29.3 &  8.7 & 27.9 & 29.2 & 29.0 & 28.7 & 28.3 \\
& pt-nl & 26.4 &  5.4 & 23.7 & 25.7 & 25.6 & 24.8 & 25.2 \\
& \textbf{Average} & \textbf{27.1} &  \textbf{8.2} & \textbf{25.1} & \textbf{26.7} & \textbf{26.4} & \textbf{25.8} & \textbf{25.6}\\
\bottomrule
\end{longtable}
}
\subsubsection{Multiway Data}
{\scriptsize
\begin{longtable}{ llrrrrrrr } 
\toprule
& Direction & Pivot (X$\rightarrow$en$\rightarrow$Y) & Baseline & Residual & Residual$+$Query \\
\midrule
\endhead
\multirow{17}{*}{\rotatebox[origin=c]{90}{\bf{Supervised}}} 
& da-en & -     &   36.3 & 36.1 & 35.0\\
& de-en & -     &   33.6 & 33.4 & 32.3\\
& es-en & -     &   40.5 & 40.1 & 39.1\\
& fi-en & -     &   30.3 & 30.1 & 29.2\\
& fr-en & -     &   37.0 & 36.8 & 35.2\\
& it-en & -     &   34.7 & 34.4 & 33.4\\
& nl-en & -     &   32.3 & 32.2 & 31.2\\
& pt-en & -     &   38.3 & 37.9 & 37.3\\
& en-da & -     &   35.7  & 35.7    & 34.8\\
& en-de & -     &   28.0  & 27.1    & 26.5\\
& en-es & -     &   42.0  & 41.4    & 41.2\\
& en-fi & -     &   21.3  & 20.8    & 20.5\\
& en-fr & -     &   37.5  & 37.2    & 36.7\\
& en-it & -     &   32.2  & 31.8    & 31.4\\
& en-nl & -     &   29.3  & 29.2    & 28.7\\
& en-pt & -     &   37.9  & 37.6    & 37.2\\
& \multicolumn{2}{l}{\textbf{Average}} 
               & \textbf{34.2} & \textbf{33.9}  & \textbf{33.1}   \\
\midrule
\multirow{57}{*}{\rotatebox[origin=c]{90}{\bf{Zero-shot}}} 
& da-de & 23.3 &    10.0 & 23.9     & 22.8\\
& da-es & 31.6 &    11.8 & 32.2     & 31.4\\
& da-fi & 17.3 &     6.2 & 17.0     & 16.4\\
& da-fr & 29.5 &    14.4 & 29.0     & 27.8\\
& da-it & 25.1 &    14.3 & 24.8     & 24.2\\
& da-nl & 25.2 &    11.2 & 25.4     & 24.1\\
& da-pt & 28.6 &    12.9 & 28.9     & 28.3\\

& de-da & 27.5 &    14.8 &  28.4    & 26.9\\
& de-es & 30.5 &    10.1 &  31.4    & 30.7\\
& de-fi & 16.2 &     5.2 &  16.2    & 15.1\\
& de-fr & 28.7 &    11.6 &  28.5    & 27.2\\
& de-it & 24.3 &    12.4 &  24.1    & 23.1\\
& de-nl & 24.6 &     9.8 &  25.4    & 24.2\\
& de-pt & 27.7 &    10.5 &  28.3    & 27.0\\

& es-da & 30.3 &    16.4 &  30.0    & 28.9\\
& es-de & 23.8 &     9.4 &  24.0    & 22.3\\
& es-fi & 18.4 &     6.3 &  17.8    & 16.9\\
& es-fr & 33.3 &    15.6 &  34.0    & 32.8\\
& es-it & 28.4 &    17.3 &  29.4    & 28.5\\
& es-nl & 25.9 &    10.7 &  26.2    & 24.7\\
& es-pt & 33.2 &    16.1 &  35.4    & 34.6\\

& fi-da & 24.9 &    13.4 &  24.6    & 24.0\\
& fi-de & 19.6 &     7.9 &  19.4    & 18.3\\
& fi-es & 27.8 &    10.3 &  27.7    & 27.2\\
& fi-fr & 26.0 &    11.5 &  25.2    & 23.9\\
& fi-it & 21.8 &    11.8 &  21.2    & 20.5\\
& fi-nl & 21.9 &     9.0 &  21.4    & 20.4\\
& fi-pt & 25.1 &    11.0 &  24.8    & 24.0\\

& fr-da & 28.3 &    14.2 &  27.5    & 26.1\\
& fr-de & 22.6 &     8.3 &  22.1    & 20.5\\
& fr-es & 33.9 &    12.1 &  35.6    & 34.4\\
& fr-fi & 17.0 &     5.3 &  16.4    & 15.3\\
& fr-it & 26.9 &    15.3 &  28.1    & 26.9\\
& fr-nl & 25.1 &     9.6 &  25.0    & 23.9\\
& fr-pt & 31.0 &    12.8 &  32.5    & 31.3\\

& it-da & 26.7 &    13.8 &  26.3    & 25.2\\
& it-de & 21.5 &     7.7 &  21.0    & 19.4\\
& it-es & 32.2 &    11.7 &  33.9    & 33.4\\
& it-fi & 15.9 &     4.7 &  15.0    & 14.8\\
& it-fr & 29.9 &    12.2 &  31.2    & 29.8\\
& it-nl & 24.1 &     9.2 &  23.6    & 22.4\\
& it-pt & 29.5 &    12.2 &  30.7    & 29.8\\

& nl-da & 26.4 &    15.4 &  27.1    & 26.1\\
& nl-de & 22.0 &     9.3 &  22.8    & 21.6\\
& nl-es & 29.3 &    10.8 &  30.2    & 29.3\\
& nl-fi & 15.8 &     5.6 &  15.5    & 14.7\\
& nl-fr & 27.7 &    12.5 &  27.7    & 26.7\\
& nl-it & 23.3 &    13.4 &  23.6    & 22.6\\
& nl-pt & 26.7 &    12.1 &  27.1    & 26.2\\

& pt-da & 29.3 &    14.3 &  28.8    & 27.7\\
& pt-de & 23.4 &     8.3 &  23.0    & 21.6\\
& pt-es & 35.8 &    14.2 &  38.0    & 37.0\\
& pt-fi & 17.6 &     5.3 &  17.0    & 16.3\\
& pt-fr & 32.6 &    13.8 &  33.4    & 32.1\\
& pt-it & 27.9 &    16.0 &  28.6    & 28.1\\
& pt-nl & 25.5 &     9.3 &  25.4    & 24.3\\

& \textbf{Average} & \textbf{25.9}
               & \textbf{11.3} & \textbf{26.1}  & \textbf{25.1}   \\

\bottomrule
\end{longtable}
}

\subsubsection{Full Data}
{\scriptsize
\begin{longtable}{ llrrrrrrr } 
\toprule
& Direction & Pivot (X$\rightarrow$en$\rightarrow$Y) & Baseline & Residual & Residual$+$Query \\
\midrule
\endhead
\multirow{17}{*}{\rotatebox[origin=c]{90}{\bf{Supervised}}} 
&da-en & - & 38.2 & 39.2 & 38.7 \\
&de-en & - & 35.7 & 36.8 & 36.4 \\
&es-en & - & 43.3 & 44.1 & 43.6 \\
&fi-en & - & 32.4 & 33.5 & 33.0 \\
&fr-en & - & 39.8 & 40.5 & 40.1 \\
&it-en & - & 37.4 & 38.4 & 38.1 \\
&nl-en & - & 33.9 & 34.9 & 34.3 \\
&pt-en & - & 41.5 & 42.3 & 41.3 \\
&en-da & - & 35.7 & 36.8 & 36.3 \\
&en-de & - & 27.9 & 29.0 & 28.4 \\
&en-es & - & 41.9 & 43.0 & 42.4 \\
&en-fi & - & 22.0 & 22.9 & 22.3 \\
&en-fr & - & 37.7 & 38.3 & 37.8 \\
&en-it & - & 32.1 & 33.1 & 32.7 \\
&en-nl & - & 29.1 & 30.4 & 29.9 \\
&en-pt & - & 38.0 & 39.1 & 38.8 \\
& \textbf{Average} &  &  \textbf{35.4} & \textbf{36.4} & \textbf{35.9} \\
\midrule
\multirow{57}{*}{\rotatebox[origin=c]{90}{\bf{Zero-shot}}} 
&da-de & 26.1 & 13.6 & 25.5 & 24.8\\
&da-es & 34.5 & 27.5 & 33.7 & 33.0\\
&da-fi & 20.0 & 15.9 & 18.9 & 17.8\\
&da-fr & 31.8 & 21.7 & 30.1 & 29.3\\
&da-it & 27.5 & 16.8 & 26.3 & 25.9\\
&da-nl & 27.5 & 10.4 & 26.2 & 26.1\\
&da-pt & 31.0 & 24.8 & 30.6 & 29.3\\
&de-da & 30.5 & 16.1 & 29.7 & 28.7\\
&de-es & 33.6 & 22.8 & 32.6 & 31.6\\
&de-fi & 19.1 & 14.0 & 18.2 & 16.0\\
&de-fr & 31.3 & 17.3 & 30.1 & 29.4\\
&de-it & 26.7 & 14.0 & 25.5 & 24.5\\
&de-nl & 27.5 & 8.5 & 26.1 & 25.4\\
&de-pt & 30.5 & 21.8 & 29.8 & 27.9\\
&es-da & 32.9 & 18.5 & 31.2 & 30.3\\
&es-de & 26.5 & 13.9 & 24.7 & 24.5\\
&es-fi & 21.0 & 15.8 & 19.6 & 18.4\\
&es-fr & 36.2 & 24.1 & 34.9 & 34.3\\
&es-it & 31.1 & 18.7 & 31.0 & 30.2\\
&es-nl & 28.4 & 11.0 & 26.7 & 26.3\\
&es-pt & 36.4 & 30.2 & 36.8 & 35.9\\
&fi-da & 27.7 & 15.4 & 25.9 & 24.9\\
&fi-de & 22.3 & 11.2 & 20.5 & 19.9\\
&fi-es & 30.7 & 22.4 & 29.3 & 28.2\\
&fi-fr & 28.8 & 17.4 & 26.1 & 24.8\\
&fi-it & 24.4 & 13.5 & 22.9 & 22.2\\
&fi-nl & 24.2 & 8.6 & 22.2 & 21.6\\
&fi-pt & 28.0 & 20.9 & 26.7 & 25.0\\
&fr-da & 30.5 & 16.8 & 29.4 & 28.5\\
&fr-de & 25.0 & 12.4 & 23.8 & 23.0\\
&fr-es & 36.8 & 27.4 & 37.4 & 36.5\\
&fr-fi & 19.6 & 14.6 & 18.5 & 16.9\\
&fr-it & 29.6 & 16.5 & 29.8 & 29.1\\
&fr-nl & 27.5 & 9.8 & 26.6 & 25.7\\
&fr-pt & 33.7 & 26.8 & 34.6 & 33.0\\
&it-da & 29.1 & 15.7 & 28.2 & 26.7\\
&it-de & 23.9 & 11.3 & 22.0 & 21.6\\
&it-es & 34.8 & 26.1 & 35.4 & 34.5\\
&it-fi & 18.3 & 13.3 & 16.5 & 15.0\\
&it-fr & 32.5 & 19.6 & 30.3 & 29.3\\
&it-nl & 26.2 & 8.9 & 25.0 & 24.0\\
&it-pt & 31.9 & 24.9 & 32.5 & 31.5\\
&nl-da & 28.7 & 16.5 & 27.9 & 27.1\\
&nl-de & 24.3 & 12.4 & 23.2 & 23.5\\
&nl-es & 31.5 & 23.6 & 31.3 & 30.4\\
&nl-fi & 17.9 & 13.7 & 17.2 & 15.4\\
&nl-fr & 29.9 & 18.6 & 29.1 & 28.2\\
&nl-it & 25.5 & 15.4 & 24.4 & 23.9\\
&nl-pt & 29.0 & 22.4 & 28.5 & 26.9\\
&pt-da & 31.7 & 17.2 & 30.2 & 29.0\\
&pt-de & 25.7 & 13.0 & 24.4 & 23.7\\
&pt-es & 38.9 & 30.8 & 39.3 & 38.3\\
&pt-fi & 20.0 & 14.8 & 19.2 & 17.5\\
&pt-fr & 35.4 & 23.1 & 34.8 & 33.9\\
&pt-it & 30.4 & 17.7 & 30.3 & 29.5\\
&pt-nl & 27.9 & 10.4 & 26.2 & 25.5\\

& \textbf{Average}  & \textbf{28.4} & \textbf{17.5} & \textbf{27.5} & \textbf{26.5}   \\
\bottomrule
\end{longtable}
}
\subsection{PMIndia}

{\scriptsize
\begin{longtable}{ llrrrrrrrr }
\toprule
& Direction & Pivot (X$\rightarrow$en$\rightarrow$Y) & Baseline & $+$vardrop & Residual & $+$vardrop & Residual$+$Query \\
\midrule
\endhead
\multirow{19}{*}{\rotatebox[origin=c]{90}{\bf{Supervised}}} 
&te-en & - & 31.2 & 30.2 & 30.7 & 30.2 & 29.7\\
&kn-en & - & 31.4 & 31.1 & 31.3 & 31.1 & 30.3\\
&ml-en & - & 28.9 & 27.9 & 28.2 & 27.9 & 27.9\\
&bn-en & - & 25.6 & 25.4 & 25.0 & 25.4 & 25.0\\
&gu-en & - & 35.3 & 34.7 & 34.7 & 34.7 & 33.5\\
&hi-en & - & 37.8 & 37.1 & 37.0 & 37.1 & 35.8\\
&mr-en & - & 29.0 & 28.7 & 28.8 & 28.7 & 28.1\\
&or-en & - & 29.0 & 28.9 & 28.9 & 28.9 & 28.3\\
&pa-en & - & 35.9 & 35.2 & 35.1 & 35.2 & 34.5\\
&en-te & - & 16.3 & 16.7 & 16.1 & 16.1 & 15.7\\
&en-kn & - & 33.4 & 33.4 & 32.6 & 32.6 & 32.2\\
&en-ml & - & 20.4 & 20.8 & 20.3 & 20.3 & 20.1\\
&en-bn & - & 22.6 & 22.5 & 21.9 & 21.9 & 21.8\\
&en-gu & - & 39.3 & 39.6 & 38.9 & 38.9 & 38.0\\
&en-hi & - & 32.5 & 32.6 & 31.5 & 31.5 & 30.8\\
&en-mr & - & 25.7 & 25.7 & 25.4 & 25.4 & 24.8\\
&en-or & - & 33.7 & 33.7 & 33.0 & 33.0 & 32.6\\
&en-pa & - & 38.8 & 38.9 & 38.0 & 38.0 & 37.1\\
& \textbf{Average} &  & \textbf{30.4} & \textbf{30.2} & \textbf{29.9} & \textbf{29.8} & \textbf{29.2} \\
\midrule
\multirow{73}{*}{\rotatebox[origin=c]{90}{\bf{Zero-shot}}} 
&te-kn & 24.9 & 0.6 & 3.2 & 2.5 & 16.3 & 1.3\\
&te-ml & 15.8 & 1.0 & 3.8 & 3.2 & 10.8 & 1.5\\
&te-bn & 18.0 & 0.7 & 2.7 & 2.0 & 12.4 & 2.0\\
&te-gu & 28.9 & 1.2 & 2.4 & 2.3 & 19.3 & 0.9\\
&te-hi & 21.6 & 0.4 & 0.5 & 1.0 & 12.8 & 1.9\\
&te-mr & 19.5 & 1.2 & 3.8 & 2.6 & 12.6 & 1.7\\
&te-or & 25.4 & 0.7 & 2.3 & 2.6 & 17.2 & 1.1\\
&te-pa & 28.2 & 0.5 & 2.1 & 2.5 & 17.9 & 0.6\\
&kn-te & 12.8 & 0.6 & 2.0 & 2.0 & 8.4 & 0.7\\
&kn-ml & 16.0 & 0.9 & 3.6 & 3.1 & 10.9 & 1.8\\
&kn-bn & 17.9 & 0.9 & 2.6 & 2.3 & 12.1 & 2.2\\
&kn-gu & 29.3 & 1.1 & 2.6 & 2.4 & 19.9 & 0.8\\
&kn-hi & 22.2 & 0.4 & 0.5 & 1.0 & 12.9 & 0.4\\
&kn-mr & 19.7 & 1.2 & 3.6 & 2.9 & 12.8 & 1.7\\
&kn-or & 25.9 & 0.6 & 2.4 & 2.6 & 17.4 & 1.4\\
&kn-pa & 29.0 & 0.5 & 2.2 & 2.7 & 18.4 & 0.7\\
&ml-te & 11.9 & 0.6 & 2.3 & 1.9 & 8.2 & 0.7\\
&ml-kn & 24.0 & 0.5 & 3.3 & 2.8 & 16.0 & 1.2\\
&ml-bn & 17.5 & 0.9 & 2.8 & 2.3 & 12.0 & 2.0\\
&ml-gu & 27.3 & 1.1 & 2.4 & 2.3 & 17.9 & 0.9\\
&ml-hi & 20.5 & 0.4 & 0.5 & 1.1 & 11.9 & 0.4\\
&ml-mr & 18.9 & 1.3 & 3.9 & 3.0 & 12.1 & 1.8\\
&ml-or & 24.8 & 0.6 & 2.3 & 2.5 & 16.3 & 1.3\\
&ml-pa & 27.0 & 0.5 & 2.0 & 2.4 & 16.7 & 0.6\\
&bn-te & 10.7 & 0.5 & 2.4 & 1.9 & 7.4 & 0.7\\
&bn-kn & 21.6 & 0.7 & 2.9 & 2.8 & 14.2 & 1.3\\
&bn-ml & 14.0 & 1.0 & 3.9 & 3.2 & 9.3 & 1.7\\
&bn-gu & 25.3 & 1.1 & 2.7 & 2.1 & 17.8 & 1.0\\
&bn-hi & 18.9 & 0.5 & 0.7 & 1.1 & 11.9 & 0.5\\
&bn-mr & 17.4 & 1.6 & 4.2 & 3.0 & 11.7 & 1.8\\
&bn-or & 23.8 & 0.9 & 3.0 & 3.1 & 16.7 & 1.3\\
&bn-pa & 25.1 & 0.6 & 2.5 & 2.5 & 16.9 & 0.6\\
&gu-te & 13.2 & 0.4 & 1.6 & 1.6 & 8.4 & 0.5\\
&gu-kn & 26.6 & 0.5 & 2.3 & 2.7 & 16.6 & 1.2\\
&gu-ml & 16.9 & 0.7 & 3.0 & 2.6 & 10.7 & 1.5\\
&gu-bn & 19.2 & 0.7 & 1.8 & 1.9 & 12.5 & 1.8\\
&gu-hi & 25.6 & 0.4 & 0.5 & 1.3 & 15.7 & 0.4\\
&gu-mr & 21.4 & 1.4 & 3.2 & 2.8 & 14.5 & 1.7\\
&gu-or & 28.1 & 0.6 & 2.0 & 2.9 & 18.1 & 1.3\\
&gu-pa & 32.0 & 0.5 & 2.3 & 2.9 & 20.6 & 0.7\\
&hi-te & 14.0 & 0.4 & 1.1 & 1.4 & 8.5 & 0.5\\
&hi-kn & 28.0 & 0.5 & 1.3 & 1.9 & 15.6 & 1.0\\
&hi-ml & 17.7 & 0.6 & 1.9 & 2.2 & 10.4 & 1.1\\
&hi-bn & 20.0 & 0.5 & 1.2 & 1.4 & 12.9 & 1.2\\
&hi-gu & 34.9 & 1.0 & 1.6 & 1.9 & 21.1 & 0.8\\
&hi-mr & 22.6 & 1.0 & 2.6 & 2.4 & 14.5 & 1.6\\
&hi-or & 30.6 & 0.5 & 1.4 & 2.0 & 18.1 & 1.0\\
&hi-pa & 35.3 & 0.4 & 1.5 & 2.3 & 21.7 & 0.6\\
&mr-te & 11.9 & 0.5 & 1.9 & 1.8 & 8.0 & 0.7\\
&mr-kn & 24.1 & 0.6 & 2.8 & 2.5 & 15.7 & 1.4\\
&mr-ml & 15.6 & 0.9 & 3.5 & 2.9 & 10.4 & 1.5\\
&mr-bn & 17.4 & 0.9 & 2.4 & 2.2 & 12.1 & 2.1\\
&mr-gu & 28.5 & 1.1 & 2.5 & 2.2 & 19.5 & 0.9\\
&mr-hi & 21.6 & 0.5 & 0.7 & 1.4 & 13.4 & 0.6\\
&mr-or & 25.3 & 0.7 & 2.2 & 2.6 & 17.0 & 1.4\\
&mr-pa & 27.6 & 0.6 & 2.2 & 2.8 & 17.0 & 0.7\\
&or-te & 11.4 & 0.6 & 2.0 & 1.8 & 8.0 & 0.6\\
&or-kn & 23.1 & 0.7 & 2.9 & 3.0 & 15.5 & 1.5\\
&or-ml & 14.9 & 0.9 & 3.5 & 3.2 & 10.3 & 1.6\\
&or-bn & 17.5 & 0.9 & 3.0 & 2.6 & 12.7 & 2.3\\
&or-gu & 28.2 & 1.3 & 3.0 & 2.7 & 19.6 & 1.0\\
&or-hi & 21.7 & 0.4 & 0.8 & 1.5 & 13.9 & 0.5\\
&or-mr & 18.7 & 1.5 & 4.2 & 3.1 & 12.3 & 1.7\\
&or-pa & 27.7 & 0.6 & 2.7 & 3.1 & 18.7 & 0.7\\
&pa-te & 12.8 & 0.5 & 1.6 & 1.8 & 8.5 & 0.5\\
&pa-kn & 26.4 & 0.6 & 2.3 & 2.6 & 16.2 & 1.1\\
&pa-ml & 16.6 & 0.8 & 3.2 & 2.7 & 10.6 & 1.5\\
&pa-bn & 18.9 & 0.8 & 2.0 & 1.9 & 12.6 & 1.9\\
&pa-gu & 32.4 & 1.4 & 2.6 & 2.6 & 22.0 & 1.0\\
&pa-hi & 26.8 & 0.5 & 0.6 & 1.5 & 17.2 & 0.4\\
&pa-mr & 21.3 & 1.4 & 3.4 & 3.0 & 13.8 & 1.7\\
&pa-or & 28.0 & 0.7 & 2.2 & 2.7 & 18.6 & 1.2\\
& \textbf{Average} & \textbf{22.1} & \textbf{0.8} & \textbf{2.4} & \textbf{2.3} & \textbf{14.3} & \textbf{1.1} \\
\bottomrule
\end{longtable}
}
\end{document}